%% file: main.tex
\documentclass[11pt]{article}
\usepackage{coling2018}
\usepackage{times}
\usepackage{url}
\usepackage{latexsym}

\usepackage[utf8]{inputenc}
\usepackage[usenames, dvipsnames]{color}
\usepackage{graphicx}
\usepackage{xcolor}
\usepackage{examples}
\usepackage{enumerate}
\usepackage{enumitem,pifont}
\usepackage{soul}
\usepackage{listings}
\usepackage{standalone}
\usepackage{multirow}
\usepackage{enumitem}
\usepackage{caption}

\lstdefinestyle{xmg}{
  keywordstyle = \color{HHUblue}\bfseries,
  commentstyle=\color{lightgray},
  literate={->}{{{\textbf{->}}}}1 {\{}{{{\textbf{\{}}}}1 {\}}{{{\textbf{\}}}}}1 {\;}{{{\textbf{;}}}}1 {|}{{{\textbf{|}}}}1 {=}{{{\textbf{=}}}}1 {[}{{{\textbf{[}}}}1 {]}{{{\textbf{]}}}}1 {<}{{{\textbf{<}}}}1 {>}{{{\textbf{>}}}}1 {!}{{{\textbf{!}}}}1 {?}{{{\textbf{?}}}}1 {*=}{{{\textbf{*=}}}}1  {'c}{{\'c}}1,%
  morekeywords={node,type,feature,include,class,import,export,declare,syn,sem,value, use, with, dims,frame,morph}
  }

\lstnewenvironment{xmg}{%
  \lstset{language=,
    numbers=left,numbersep=8pt,numberstyle=\color{lightgray},
    alsoletter={-},%
    basicstyle=\small\ttfamily,%
    xleftmargin=0.7cm,framexleftmargin=12pt,%
    framerule=0.5mm,rulecolor=\color{lightgray},%
    framesep=0pt,%
    escapeinside={|\%}{\%|},%
    commentstyle=\color{lightgray},
    literate={->}{{{\textbf{->}}}}1 {<-}{{{\textbf{<-}}}}1 {\{}{{{\textbf{\{}}}}1 {\}}{{{\textbf{\}}}}}1 {\;}{{{\textbf{;}}}}1 {|}{{{\textbf{|}}}}1 {=}{{{\textbf{=}}}}1 {[}{{{\textbf{[}}}}1 {]}{{{\textbf{]}}}}1 {<}{{{\textbf{<}}}}1 {>}{{{\textbf{>}}}}1 {!}{{{\textbf{!}}}}1 {?}{{{\textbf{?}}}}1 {*=}{{{\textbf{*=}}}}1,%
    morekeywords={node,type,feature,include,class,import,export,declare,syn,frame-types,frame-constraints,frame,iface,morph,morpho,lemma,value, use, with, dims}}}{}

\newcommand{\ixmg}{%
  \lstinline[language=,keepspaces,%
      basicstyle=\small\ttfamily,%
      literate={->}{{{\textbf{->}}}}1 {<-}{{{\textbf{<-}}}}1 {\{}{{{\textbf{\{}}}}1 {\}}{{{\textbf{\}}}}}1 {\;}{{{\textbf{;}}}}1 {|}{{{\textbf{|}}}}1 {=}{{{\textbf{=}}}}1 {[}{{{\textbf{[}}}}1 {]}{{{\textbf{]}}}}1 {<}{{{\textbf{<}}}}1 {>}{{{\textbf{>}}}}1 {!}{{{\textbf{!}}}}1 {?}{{{\textbf{?}}}}1 {*=}{{{\textbf{*=}}}}1,%
      morekeywords={node,type,feature,include,class,import,export,declare,{frame-types},syn,frame,iface,morph,morpho,lemma,value, use, with, dims}
      ]}

\newcommand{\lex}[1]{\textbf{#1}}  
\newcommand{\ile}[1]{\textsl{#1}} 
\newcommand{\lit}[1]{`#1'} 
\newcommand{\idio}[1]{`#1'}  
\newcommand{\exlit}[2]{\ile{#1}~\lit{#2}} 
\newcommand{\exidio}[2]{\ile{#1}~\idio{#2}} 
\newcommand{\litidio}[2]{\lit{#1}$ \Rightarrow $\idio{#2}} 
\newcommand{\exlitidio}[3]{\ile{#1}~\lit{#2}$\Rightarrow$\idio{#3}} 

\title{Object-oriented lexical encoding of multiword expressions:\\ Short and sweet}

\author{Agata Savary \\
  University of Tours\\
  France \\
  {\tt {\small first.last@univ-tours.fr}} \\\And
  Simon Petitjean  \\
  Heinrich-Heine-Universität Düsseldorf\\
  Germany \\
  {\tt {\small last@phil.uni-duesseldorf.de}} \\\AND
  Timm Lichte  \\
  University of Tübingen\\
  Germany \\
  {\tt {\small first.last@uni-tuebingen.de}} \And
  Laura Kallmeyer \\
  Heinrich-Heine-Universität Düsseldorf\\
  Germany \\
  {\tt {\small last@phil.uni-duesseldorf.de}} \\\AND
  Jakub Waszczuk  \\
  Heinrich-Heine-Universität Düsseldorf\\
  Germany \\
  {\tt {\small first.last@phil.uni-duesseldorf.de}} \\
}

\date{}

\begin{document}
\maketitle
\begin{abstract}
Multiword expressions (MWEs) exhibit both regular and idiosyncratic properties. Their idiosyncrasy requires lexical encoding in parallel with their component words. Their (at times intricate) regularity, on the other hand, calls for means of flexible factorization  
to avoid redundant descriptions of shared properties. However, so far,  non-redundant general-purpose lexical encoding of MWEs has not received a satisfactory solution. We offer a proof of concept that this challenge might be effectively addressed within eXtensible MetaGrammar (XMG), an object-oriented metagrammar framework. We first make an existing metagrammatical resource, the FrenchTAG grammar, MWE-aware. We then evaluate the factorization gain during incremental implementation with XMG on a dataset extracted from an MWE-annotated reference corpus.
%
%
%

\end{abstract}

%
%
    %
    %
    %
    %
    %
    %

\section{Introduction} 
\label{sec:intro}
\input{intro}

\section{Related work}
\label{sec:related}
\input{related}

\section{From a metagrammar to parsing} 
\label{sec:metagrammar}
\input{metagrammar}

\section{FrenchTAG: A French XMG metagrammar} 
\label{sec:frenchtag}
\input{frenchtag}

\section{Enriching a metagrammar with MWEs} 
\label{sec:mwes}
\input{mwes}

\section{Evaluation}
\label{sec:eval}
\input{eval}
\section{Conclusions and future work} 
\label{sec:conc}
\input{conc}




\bibliography{biblio}
\bibliographystyle{acl}

\section*{Appendix: sentences from the evaluation corpus}

\begin{small}
\textbf{DEV-S}:
\begin{examples}
\vspace{-2mm}\item \exlitidio{la unification \lex{avait lieu} de être}{the unification has place of be}{the unification has good reasons to take place}
\vspace{-3mm}\item \exlitidio{une évolution \lex{avait} alors \lex{lieu}}{an evolution has then place}{then an evolution took place}
\vspace{-3mm}\item \exlitidio{ils \lex{faisaient appel} à des charpentiers}{they made appeal to carpenters}{they appealed to carpenters}
\vspace{-3mm}\item \exlitidio{les pilotes \lex{faisaient appel}}{the pilots made appeal}{the pilots appealed}
\vspace{-3mm}\item \exlitidio{je \lex{faisais appel} à eux}{I did appeal to them}{I appealed to them}
\vspace{-3mm}\item \exlitidio{ils \lex{faisaient appel}}{they did appeal}{they appeled}
\vspace{-3mm}\item \exlitidio{les commandants \lex{faisaient} certainement \lex{face} à un dilemme}{the commanders made certainly face to a dilemma}{the commanders certainly had to deal with a dilemma}
\vspace{-3mm}\item \exlitidio{elles \lex{faisaient le objet} de un ordre}{they did the object of an order}{they were the subject of an order}
\vspace{-3mm}\item \exlitidio{ils \lex{feront le objet} d'une campagne}{they will do the object of a campaign}{they will be the object of a campaign}
\vspace{-3mm}\item \exlitidio{le service \lex{faisait} alors \lex{le objet} de les critiques}{the service did then the object of criticisms}{the service was subject to criticism}
\vspace{-3mm}\item \exlitidio{elle \lex{faisait partie} de son territoire}{they did part of its territory}{they were part of its territory}
\vspace{-3mm}\item \exlitidio{les célébrités ont \lex{fait partie} de le casting}{the celebrities did part of the casting}{the celebrities took part of the casting}
\vspace{-3mm}\item \exlitidio{\lex{il faut} le dire}{it should it say}{it should be said}
\vspace{-3mm}\item \exlitidio{\lex{il faut} choisir un nom}{it should choose a name}{one should choose a name}
\vspace{-3mm}\item \exlitidio{\lex{il se agit} de avoir une garantie}{it itself acts of have a guarantee}{the point is to have a guarantee}
\vspace{-3mm}\item \exlitidio{\lex{il se agit} certainement de un organe}{it itself acts certainly of an organ}{an organ is certainly concerned}
\vspace{-3mm}\item \exlitidio{\lex{il se agit} de le tube}{it itself acts of the hit}{the hit is concerned}
\vspace{-3mm}\item \exlitidio{\lex{il se agissait} alors de colorier}{it itself acts of color}{coloring is needed}
\vspace{-3mm}\item \exlitidio{\lex{il y a} de la concurrence}{it there has of the competition}{there is competition}
\vspace{-3mm}\item\label{ex:dev:mis-en-examen} \exlitidio{\lex{mis en examen}, Jean a été suspendu}{put under investigation, Jean has been suspended}{Jean was suspended after having been investigated}
\vspace{-3mm}\item \exlitidio{Jean est \lex{mis en examen}}{Jean is put in exam}{Jean is put under investigation}
\vspace{-3mm}\item \exlitidio{les personnes qui sont \lex{mises en examen}}{the persons who are put in exam}{the persons who are put under investigation}
\vspace{-3mm}\item\label{ex:dev:porter-nom-1} \exlitidio{le saint dont elle \lex{porte} le \lex{nom}}{the saint whose she bears the name}{the saint whose name she bears}
\vspace{-3mm}\item\label{ex:dev:porter-nom-2} \exlitidio{Paris qui \lex{porte} aujourd'hui son \lex{nom}}{Paris which bears today his name}{Paris which bears its name today}
\vspace{-3mm}\item\label{ex:dev:se-faire-adj} \exlitidio{les visions \lex{se font} intenses}{the visions make themselves intensive}{the visions become intensive}
\vspace{-3mm}\item \exlitidio{son tombeau \lex{se trouve} à Paris}{his tomb itself finds here}{his tomb is here}
\end{examples}

\noindent\textbf{TEST-S}:
\begin{examples}
\vspace{-2mm}\item \exlitidio{elle \lex{avait lieu} ici}{it had place here}{it took place here}
\vspace{-3mm}\item \exlitidio{les championnats \lex{ont lieu} ici demain}{the championships have place here tomorrow}{the championships take place here tomorrow}
\vspace{-3mm}\item \exlitidio{les figurines qui se \lex{font face}}{figures which themselves do face}{figures which face each other}
\vspace{-3mm}\item \exlitidio{la charte qui doit \lex{faire face} à le impérialisme}{the charter which must do face to the imperialism}{the charter which must face the imperialism}
\vspace{-3mm}\item \exlitidio{un buste qui \lex{fait face} à le spectateur}{a bust which does face to the spectator}{a bust which faces to the spectator}
\vspace{-3mm}\item \exlitidio{le futur \lex{fait le objet} de un débat}{the future make the object of a debate}{the future is subject to debate}
\vspace{-3mm}\item \exlitidio{elle \lex{fait partie} de le groupe}{she makes part of a group}{she is part of a group} 
\vspace{-3mm}\item \exlitidio{un monument qui \lex{fait partie} de les sites}{a monument which makes part of the sites}{a monument which is part of the sites}
\vspace{-3mm}\item \exlitidio{\lex{il} en \lex{faut}}{it thereof needs}{it is needed}
\vspace{-3mm}\item \exlitidio{\lex{il} la \lex{faut}}{it her needs}{she is needed} 
\vspace{-3mm}\item \exlitidio{\lex{il y a} contrôle}{it there has control}{there is control}
\vspace{-3mm}\item \exidio{le \lex{rôle} qui est \lex{joué} par le personnage}{the role which is played by the character}
\vspace{-3mm}\item \exlitidio{la religion \lex{joue} un \lex{rôle} majeur}{teh religion plays a role major}{religion plays a major role}
\vspace{-3mm}\item \exlitidio{des macrophytes \lex{jouent} le \lex{rôle} de barrière}{macrophytes play the role of barrier}{macrophytes play the role of a barrier}
\vspace{-3mm}\item \exlitidio{il \lex{joue} un \lex{rôle} important}{he plays a role important}{he plays an important role}
\vspace{-3mm}\item \exlitidio{un policier était \lex{mis en examen} ici}{a policeman was put in exam}{the policement was put under invesitgation}
\vspace{-3mm}\item \exlitidio{les habitants veulent y \lex{mettre fin}}{the inhabitants want there put end}{the inhabitants want to put an end to it}
\vspace{-3mm}\item \exlitidio{un dialogue qui \lex{met fin} à la occupation}{the dialogue which puts end to the occupation}{the dialogue which puts an end to the occupation}
\vspace{-3mm}\item \exlitidio{il fait \lex{mettre fin} à cette pratique}{he makes put end to this practice}{he makes the practice be stopped}
\vspace{-3mm}\item \exlitidio{il décide de \lex{mettre fin} à cette anarchie}{he decides to put end to this anarchy}{he decides to put an end to this anarchy}
\vspace{-3mm}\item\label{ex:test:porter-nom-1} \exlitidio{il verra le puits \lex{porter} son \lex{nom}}{he will-see the well bear his name}{the well will bear his name}
\vspace{-3mm}\item\label{ex:test:porter-nom-2} \exidio{la école \lex{porte} son \lex{nom} maintenant}{the school bears his name now}
\vspace{-3mm}\item \exlitidio{Jean \lex{se fait} tuer par Marie}{Jean himself makes kill by Mary}{Jean is killed by Mary}
\vspace{-3mm}\item\label{ex:test:se-trouver-adj} \exlitidio{le terrain \lex{se trouve} pris sous le feu}{the site itself finds taken under the fire}{the site finds itself under fire}
\vspace{-3mm}\item \exlitidio{Jean \lex{se trouve} ici}{Jean himself finds here}{Jean is here}
\vspace{-3mm}\item \exlitidio{il est incroyable de \lex{se trouver} ici}{it is incredible of oneself find here}{it is incredible to be here}
\end{examples}
\end{small}

\end{document}

%% file: intro.tex

Multiword expressions are combinations of words which encompass heterogeneous linguistic objects such as idioms (IDs: \ile{to \textbf{pull} one's \textbf{leg}}), compounds (\ile{a \textbf{hot dog}}), light verb constructions (LVCs: \ile{to \textbf{pay} a \textbf{visit}}), inherently reflexive verbs (IRVs: \exlitidio{\lex{s}'\lex{apercevoir}}{self
perceive}{realize} in French), rhetorical figures (\ile{\lex{as busy as a bee}}), or named entities (\ile{the \textbf{Sea of Tranquility}}). Their most pervasive and challenging feature is their non-compositional semantics, i.e. the fact that their meaning cannot be deduced from the meanings of their
components, and from their syntactic structures, in a way deemed regular for the given language. For this reason, as well as because of their pervasiveness in texts, MWEs constitute a major challenge in semantically oriented NLP applications. 

But MWEs also exhibit unexpected behavior on other levels of linguistic analysis including the lexical, morphological and syntactic ones. These properties can be \emph{defective} or \emph{restrictive} \cite{Lichteetal18}. A defective property excludes a literal interpretation of the MWE, e.g. \ile{a \textbf{cross}-\textbf{roads}} cannot be understood literally because of the lack of number agreement between the determiner and the head noun. A restrictive property reduces the number of possible surface
realizations of the MWE with respect to the literal reading. For instance in 
in example (\ref{ex:en:cross:fingers}), the possessive determiner has to agree with the subject, otherwise the expression can only be understood literally as in \#\ile{John crossed her fingers}.\footnote{The hash symbol \# signals the loss of the idiomatic reading. Lexicalized components of a MWE, i.e. those always realized by the same lexemes, are marked in boldface.} Since defective and restrictive properties help distinguish literal from idiomatic readings of MWEs, their description and processing are important both for linguistic modeling and for NLP applications, including MWE identification \cite{constant:halshs-01665254}.

\begin{examples}
\vspace{-1.5mm}\item \label{ex:en:break-mug} \ile{John broke \underline{my} mug}
\vspace{-3mm}\item \label{ex:en:break-fall} \exidio{John \textbf{broke} \underline{his}/\underline{our} \textbf{fall}}{John made his/our fall less forceful}
\vspace{-3mm}\item \label{ex:en:cross:fingers} \exidio{John \textbf{crossed} \underline{his} \textbf{fingers}}{John hoped for good luck}
\vspace{-3mm}\item \label{ex:en:hold:tongue} \exidio{John \textbf{held} \underline{his} \textbf{tongue}}{John refrained from expressing his view}
\vspace{-1.5mm}
\end{examples}

When characterizing MWEs, some authors \cite{Gregoire10,prz:etal:14b} oppose the \emph{regular} behavior of ``free'' phrases (i.e. those obeying the rules of a ``regular'' grammar), like (\ref{ex:en:break-mug}), to the \emph{idiosyncratic} behavior of MWEs, like (\ref{ex:en:break-fall})--(\ref{ex:en:hold:tongue}). Some others point out that regularity is a matter of scale rather than a binary phenomenon \cite{GastonGross88,hamw:2015-cr,Lichteetal18}. We take the latter stand, and extend it by assuming that the degree of regularity is a feature of linguistic properties on the one hand, and of MWEs on the other hand. Firstly, the more (resp. less) objects share a certain property, the more it is regular (resp. idiosyncratic). For instance, allowing a possessive determiner in a Verb-Det-Noun construction is more regular than imposing that it agrees with the subject, because the former applies to (\ref{ex:en:break-mug})--(\ref{ex:en:hold:tongue}), while the latter is limited to (\ref{ex:en:cross:fingers})--(\ref{ex:en:hold:tongue}). Still the latter is not fully irregular since it is shared by many expressions.
%
Secondly, in (\ref{ex:en:cross:fingers}), while the direct object of the verb \ile{to cross} is lexicalized (has to be realized by the lexeme \ile{finger}), the subject is not. While the noun does not admit adjectival modifiers (\#\ile{He crossed his long fingers.}), passivization is allowed (\ile{\lex{fingers crossed}}). While the noun has to occur in plural, the verb can be inflected freely, etc. Thus, this MWE combines more regular properties (e.g. a free subject) with more idiosyncratic ones (e.g. a lexically and morphologically fixed object).


Because MWEs exhibit (more or less) idiosyncratic properties, their modeling has to include lexical encoding, i.e. 
MWEs should become separate lexical entries, additionally to their single-word components. The main challenge is then to account for the irregularity of a MWE, while avoiding redundancy, i.e. repeated description of common properties. For instance, the subject-possessive agreement is shared by (\ref{ex:en:cross:fingers})--(\ref{ex:en:hold:tongue}) and many other MWEs, so its formalization should preferably be done only once, rather than repeatedly for each MWE lexicon entry.
As shown is Sec.~\ref{sec:related}, no previous work seems to have addressed this challenge in a satisfactory way. 

In this paper, we aim at providing a proof of concept that non-redundant lexical encoding of MWEs can be effectively achieved in an object-oriented metagrammar-based approach. We use XMG \cite{DBLP:journals/coling/CrabbeDGRP13,DBLP:conf/lacl/PetitjeanD016}, a declarative 
constraint-based description language in which more or less regular tree structures are modeled via a hierarchy of classes.  
%
We test our proposal on French. We accommodate FrenchTAG \cite{Crabbe:2005}, a pre-existing XMG resource which implements a large fragment of a reference grammar of French \cite{Abeille:2002}. We show how FrenchTAG can be adapted and extended so as to accommodate a small subset of verbal MWEs (VMWEs) of different syntactic structures and of varying degrees of syntactic flexibility. We evaluate the proposal on a dataset based on the PARSEME corpus of VMWEs \cite{Savaryatal:forth}. The experiment shows that adding MWE descriptions to a general grammar can be done elegantly by introducing interface constraints in pre-existing classes (to account for restrictive properties), and by adding some new classes (to account for defective properties and for various syntactic structures of lexicalized verbal arguments).

The paper is organized as follows. We discuss the state of the art in lexical encoding of MWEs in computational lexicons and grammars (Sec.~\ref{sec:related}). We 
introduce our formalisms and tools (Sec.~\ref{sec:metagrammar}). We explain the methodology of making the original metagrammar MWE-aware (Sec.~\ref{sec:mwes}). We discuss the evaluation protocol and results (Sec.~\ref{sec:eval}). Finally, we conclude and give directions for future work (Sec.~\ref{sec:conc}).


%% file: related.tex
Lexical encoding of MWEs has a long linguistic tradition, notably in French with Gross \shortcite{Gross:1986:LRC:991365.991367} and Mel'{\v{c}}uk \shortcite{Melcuketal88}. They assume that units of meaning are located at the level of elementary sentences 
rather than of words, and MWEs, especially verbal, are special instances of predicates in which some arguments are lexicalized. 
Those works paved the way towards systematic syntactic description of MWEs, but were not directly applicable to NLP due to insufficient formalization \cite{ConTol10}. 

With the growing understanding of the challenges which MWEs pose to NLP, a large number of NLP-dedicated MWE lexicons have been created for many languages \cite{LOSNEGAARD16.718}, some of which account for 
selected morpho-syntactic properties. They can be classified notably along two axes: 
\begin{enumerate}[label={Axis \arabic*},leftmargin=*]
\vspace{-2mm}\item Formalization of the lexicon-grammar interaction. \label{axis1}
\vspace{-2mm}\item Existence of factorization mechanisms. \label{axis2}
\vspace{-2mm}\end{enumerate}

\ref{axis1} introduces a division between works which account for \emph{continuous} MWEs only (i.e. those whose components are adjacent in text) and possibly \emph{discontinuous} ones. In the former case \cite{Savary08}, finite-state-related formalisms, possibly enriched with unification, are often used \cite{KarKapZae92,BreSegVal96,OflCetSay04,Silberztein05} 
since only local phenomena need to be covered, and there is no need to account for the grammar of a language in a comprehensive way.

Conversely, the description of discontinuous MWE, most prominently of VMWEs, usually calls for more or less explicit reference to a full-fledged grammar, because of interactions between MWEs and external elements. For instance, the MWE in (\ref{ex:en:break-fall}) has a compulsory but non-lexicalized modifier of the noun \ile{fall}, which can be realized by syntactically complex nominal phrases (\ile{John \textbf{broke} \underline{his secretly adored office mate's} \textbf{fall}}). Such long-distance dependencies have been covered with two objectives in mind: (i) theory-independence and (ii) integration with computational grammars. 
 Firstly, it was postulated that MWE encoding, which is a labor intensive task, calls for a theory-neutral framework \cite{Gregoire10,prz:etal:14b,McShaneetal15}. 
These works assume the existence of general grammar rules (of the language under study), whose observance a native lexicographer is able to verify. The description of a MWE is then done in such a way that only its idiosyncratic properties (i.e. those not conforming to the regular grammar) are encoded, while the regular ones are assumed implicitly. Although these lexicons suffer from insufficient formalization 
\cite{Lichteetal18}, 
they could be successfully applied to parsing after ad hoc conversion to particular grammar formalisms \cite{pat:15}.
%
Secondly, a range of computational grammars accommodate some types of MWEs directly in their lexicons. 
In Head-driven Phrase Structure Grammar (HPSG), \newcite{Sagetal02}, \newcite{Copestakeetal02}, and \newcite{Villavicencioetal04} represent decomposable English MWEs (\ile{to \lex{spill} the \lex{beans}}) by paraphrasing (\ile{spill} $\Rightarrow$ \idio{reveal}, \ile{the beans} $\Rightarrow$ \idio{a secret})
 and MWEs with opaque semantics (\ile{to \lex{kick the bucket}}) by separate semantic predicates. \newcite{bhf:2015} additionally focus on co-indexation mechanisms needed to represent possessed idioms such as (\ref{ex:en:cross:fingers})--(\ref{ex:en:hold:tongue}). Finally, \newcite{hamw:2015-cr} adopt a largely compositional analysis of MWEs in Hebrew by introducing dedicated lexicon entries for each lexicalized component of each MWE. In Lexical Functional Grammar (LFG), \newcite{Attia06} parses Arabic continuous semi-fixed MWEs (\ile{\lex{traffic light}}) as single tokens, while syntactically compositional but semantically non-compositional MWEs (\litidio{fiery bike}{motorbike}) are handled by the grammar via lexical selection rules, similarly to the HPSG approaches. 
 A formally very different account is found in Lexicalized Tree-Adjoining Grammar (LTAG). Since the elementary structures of LTAG, the elementary trees, correspond to an ``extended domain of locality'', even the surface structure of discontinuous MWEs can be directly represented within the lexicon \cite{AbeilleSchabes89,Abeille:Schabes:96,Vaidya-etal:2014,LichteKallmeyer:2016}. This sort of approach is therefore most similar to the words-with-spaces approaches in HPSG and LFG, yet making available more structure and including slots rather than spaces.  
As a conclusion, there exist, on the one hand, generic lexical MWE resources which suffer from the lack of sufficient formalization, and, on the other hand, perfectly formalized solutions but restricted to particular grammar formalisms.

Along \ref{axis2}, the challenge to address is the proliferation of idiosyncrasy profiles of MWEs. Some MWE lexicons do not introduce generalization of the MWE behavior \cite{al-haj:etal:2014}. Some do it via inflection codes \cite{Savary09}, equivalence classes \cite{Gregoire10}, macros \cite{prz:etal:14b}, or type hierarchies \cite{hamw:2015-cr}, with a limited degree of recursiveness. 
The metagrammatical approach by \newcite{Jacquemin01} addresses the morphological, syntactic and semantic variation of French MWEs in a factorized way. There, canonical forms of MWEs are represented as fully lexicalized CFG-like rules with feature structures and unification, while MWE variants are covered by metarules, but the proposal is restricted to continuous terminological compounds. 

In view of this state of the art, it seems that a non-redundant lexical encoding of MWEs, which would account for both continuous and discontinuous MWEs, as well as their scale-wise regularity, has not yet received a satisfactory solution.
The goal of this paper is to 
take steps towards such a solution in a metagrammatical (i.e. relatively theory-independent) framework. We focus on lexical and morpho-syntactic properties of MWEs, and present a proof of concept in the context of FTAG.






%% file: metagrammar.tex


The framework of eXtensible MetaGrammar (XMG) \cite{DBLP:journals/coling/CrabbeDGRP13,DBLP:conf/lacl/PetitjeanD016} provides description languages and dedicated compilers for generating a wide range of linguistic resources.\footnote{\url{http://dokufarm.phil.hhu.de/xmg/}} 
%
Descriptions are organized into \emph{classes}, 
as in object-oriented programming. Similarly, classes have encapsulated name spaces and inheritance relations may hold between them. The crucial elements of a class are \emph{dimensions}. They can be equipped with specific description languages and are compiled independently, thereby enabling the grammar writer to treat the levels of linguistic information separately.

In this paper, we use the \ixmg{<syn>} and \ixmg{<iface>} dimensions. \ixmg{<syn>} holds tree constraints that express dominance and precedence relations among nodes. Nodes may carry (untyped) feature structures. \ixmg{<iface>} is an interface dimension where (non-typed) feature structures are used to share information between other dimensions and classes.\footnote{An alternative notation for the \ixmg{<iface>} dimension uses the \ixmg{*=} operator as shown in the right column of Fig.~\ref{fig:mwe-classes}.} The general structure of a class is given on the left of Listing \ref{fig:xmg-classes}.
\begin{figure}[ht]
\begin{minipage}[c]{0.48\textwidth}
\begin{xmg}
class classname
import someOtherClass[]
export ?someVariable
declare ?someVariable {
  <syn>{ ... };
  <iface> { ... } }
\end{xmg}
\end{minipage}
\begin{minipage}[c]{0.48\textwidth}
\begin{xmg}
class CanSubject
export ?VN 
declare ?VN {
  <syn>{node [cat=s]{
          node [cat=n]
	  node ?VN[cat=vn]} }}
\end{xmg}
\end{minipage}
\captionof{lstlisting}{Structure of XMG classes (left) and a \ixmg{<syn>} dimension example (right).}
\label{fig:xmg-classes}
\end{figure}

The more concrete class CanSubject on the right of Listing \ref{fig:xmg-classes} describes the canonical position of a subject N on the left of the verbal nucleus VN. 
The corresponding tree fragment is given in Fig.~\ref{fig:ftag-tree-fragments}(a). Note that \ixmg{import}, \ixmg{export}, and \ixmg{declare} behave similarly to the corresponding constructs in object-oriented programming. The \ixmg{;} and \ixmg{|} operators expresses conjunction and disjunction of 
statements, respectively. 

While XMG comes with compilers (implemented as constraint solvers) for several syntactic formalisms, we focus on LTAG, due to its extended domain of locality enabling seamless representation of MWEs. We use LTAG grammars compiled from XMG metagrammars as input to the TuLiPA parser \cite{Parmentier:etal:08,Arps:Petitjean:18}, applied in the experiments reported on in Sec.~\ref{sec:eval}.   


%% file: frenchtag.tex
FrenchTAG \cite{Crabbe:2005} is a syntactic XMG implementation of the reference grammar of French by \newcite{Abeille:2002}.\footnote{FrenchTAG later evolved into SEMTAG with a unification-based compositional semantic dimension \cite{DBLP:conf/coling/Gardent08}.} It contains 285 XMG classes\footnote{Only 246 of them occur in the FrenchTAG version adapted to the current version of XMG.}, including 96 families (sets of trees assigned to lexemes), which compile into 9045 TAG trees. 
Its main focus are verbs. It defines about 40 verbal subcategorisation frames, and -- for each frame -- the allowed diatheses (active, passive, middle, reflexive, impersonal, etc.) and argument realizations (canonical, clitic, extracted, omitted, etc.). Listing \ref{fig:ftag-classes} shows an extract of the classes describing transitive verbs like \exlit{ouvrir}{open}. Invoking a class $C1$ within class $C2$ means that $C2$ inherits $C1$'s descriptions. 
The \texttt{Subject} class is a disjunction of various realizations of a subject: canonical (\exlit{Jean}{Jean}), clitic (\exlit{il}{he.\textsc{subj}}), relative (\exlit{Jean qui ouvre la porte}{Jean who opens  the door}), etc. Similarly, a direct \texttt{Object} can be canonical (\exlit{la porte}{the door}), clitic (\exlit{la}{it.\textsc{obj}}), relative (\exlit{la porte que Jean ouvre}{the door which Jean opens}), reflexive (\exlit{Jean s'ouvre}{Jean opens himself}) etc. Then, the \texttt{dian0Vn1Active} class combines any realization of a subject and an object with a verb in active voice. 
Class \texttt{n0Vn1} is a disjunction of diatheses: active (\exlit{Jean ouvre la porte}{Jean opens the door}), passive (\exlit{la porte est ouverte par Jean}{the door is opened by Jean}), short passive (\exlit{la porte est ouverte}{the door is opened}), etc. 
Class \texttt{n0ClV} represents inherently reflexive verbs discussed in Sec.~\ref{sec:mwes}. 

\begin{figure}[ht]
  \input{ftag-classes.tex}
  \captionof{lstlisting}{FrenchTAG classes for the realizations of a subject, a direct object, an active diathesis, all diatheses of transitive verbs, and of an inherently reflexive verb.}
\label{fig:ftag-classes}
\end{figure}

\begin{figure}[ht]
\setlength{\tabcolsep}{1.2mm}
\begin{tabular}{cccccc|ccc}
\includegraphics[scale=1]{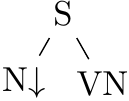} &
\includegraphics[scale=1]{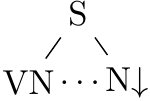} &
\includegraphics[scale=1]{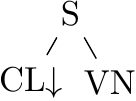} &
\includegraphics[scale=1]{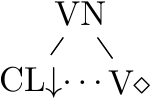} &
\includegraphics[scale=1]{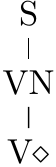} &
\includegraphics[scale=1]{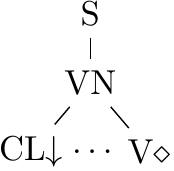} &
\includegraphics[scale=1]{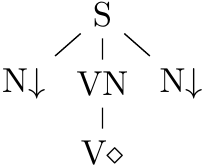} &
\includegraphics[scale=1]{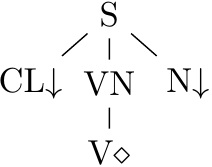} &
\includegraphics[scale=1]{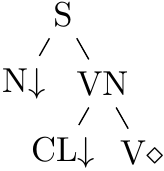} 
\\
(a) & (b) & (c) & (d) & (e) & (f) & (g) & (h) & (i)
\end{tabular}
\caption{Tree fragments described by the XMG classes from Listing \ref{fig:ftag-classes}: canonical subject (a) and object (b), clitic subject (c) and object (d), active (e) and reflexive (f) verb morphology. The corresponding LTAG trees are shown in (g--i). Feature structures are omitted. The dots in (b), (d) and (f) represent a possibly non-immediate precedence of nodes.}
\label{fig:ftag-tree-fragments}
\end{figure}

\begin{figure}[ht]
\includegraphics[scale=0.9]{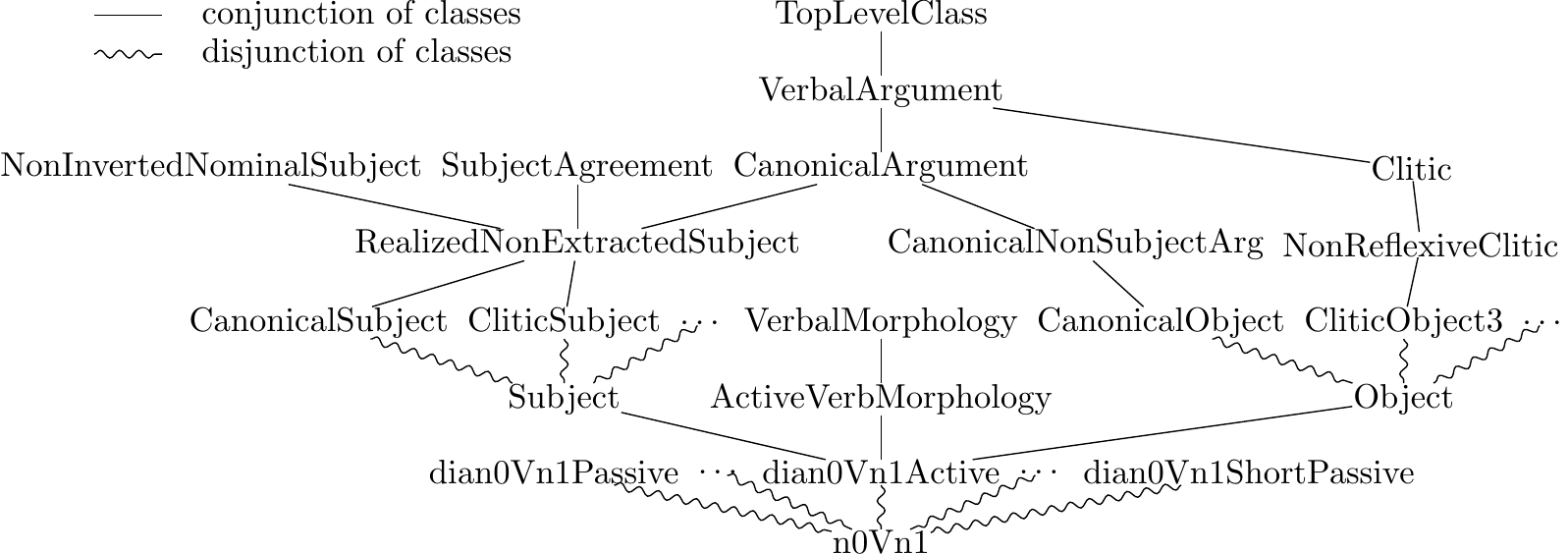} 
\caption{Extract of the XMG class hierarchy in the original FrenchTAG metagrammar.}
\label{fig:hierarchy-1}
\end{figure}

Each class describes a more or less abstract set of tree fragments, which can be made more specific by adding constraints in the inheriting classes. For instance, \texttt{CanonicalSubject} inherited by \texttt{Subject} corresponds to the tree in  Fig.~\ref{fig:ftag-tree-fragments}(a). In case of class conjunction, tree fragments of the inherited classes are combined in that shared nodes (represented by global variables, invisible here) are unified. Here, \texttt{dian0Vn1Active} yields several combinations of tree fragments (a)--(f), including tree (i) (\exlit{Jean l'ouvre}{John opens it}) obtained from (a), (d) and (e) unified along the verbal spine \texttt{S}$\rightarrow$\texttt{VN}$\rightarrow$\texttt{V}. Compiling an XMG metagrammar $M$ into an LTAG boils down to finding minimal tree models which fulfill constraints expressed in $M$. Here, the tree fragment (d) imposes that the clitic \texttt{CL} precedes the verb \texttt{V} possibly indirectly (which is signaled by dots between the sibling nodes). But since no other fragment imposes a third node between the \texttt{CL} and \texttt{V}, the minimal model is the one with a direct precedence. 
An extract of the FrenchTAG class hierarchy, with some classes from Listing \ref{fig:ftag-classes}, is shown in Fig.~\ref{fig:hierarchy-1}.


FrenchTAG, following the XTAG \cite{XTAG2001} architecture, separates the grammatical description from the lexicon. The latter covers lemmas and their inflected forms, as shown in Listing \ref{fig:ftag-lemmas}. Families assigned to the lemmas refer to grammar classes, i.e. sets of tree fragments. During parsing, the grammar is anchored with the lexicon, i.e. lexicon entries are linked with anchor nodes (marked with $\diamond$) of the grammar trees, provided that \emph{anchoring constraints} are fulfilled. The latter are based on unification of feature structures (FS) of two types: FS attached to tree nodes (neglected here for brevity) and the so-called \emph{interface FSs} discussed in the following section.

\begin{figure}[ht]
\input{ftag-lexicon.tex}
\end{minipage}\hfill
\captionof{lstlisting}{Extract of the FrenchTAG lexicon with lemmas (first 3 columns) and inflected forms (last column). Morphological features are omitted.}
\label{fig:ftag-lemmas}
\end{figure}



%% file: ftag-classes.tex
\begin{minipage}[c]{0.48\textwidth}
\begin{xmg} [caption={Caption}]
class Subject {
  CanonicalSubject[]
  | CliticSubject[]
  | RelativeSubject[]
class Object {
  CanonicalObject[]
  | CliticObjectII[]
  | RelativeObject[]
  | reflexiveAccusative[]
\end{xmg}
\end{minipage}\hfill
\begin{minipage}[c]{0.48\textwidth}
\begin{xmg}
class dian0Vn1Active {
  Subject[];
  activeVerbMorphology[];
  Object[] }
class n0Vn1 {
  dian0Vn1Active[]
  | dian0Vn1Passive[]
  | dian0Vn1ShortPassive[]
class n0ClV{
  Subject[];
  reflexiveVerbMorphology[] }
\end{xmg}
\end{minipage}

%% file: ftag-lexicon.tex
\begin{minipage}[c]{0.25\textwidth}
\begin{xmg}
class LemJean {
<lemma> {
 entry <- "Jean";
 cat <- n;
 fam <- propn}}
class LemIl {
<lemma> {
 entry <- "il";
 cat <- cl;
 fam <- CliticT}}
\end{xmg}
\end{minipage}\hfill
\begin{minipage}[c]{0.25\textwidth}
\begin{xmg}
class LemOuvrir {
<lemma> {
 entry <-"ouvrir";
 cat <- v;
 fam <- n0Vn1}}
class LemTaire {
<lemma> {
 entry <-"taire";
 cat <- v;
 fam <- n0ClV}}
\end{xmg}
\end{minipage}\hfill
\begin{minipage}[c]{0.25\textwidth}
\begin{xmg}
class LemPorte{
<lemma> {
 entry <- "porte";
 cat <- n;
 fam <- noun }}
class LemLe {
<lemma> {
 entry <- "le";
 cat <- d;
 fam <- stddet }}
\end{xmg}
\end{minipage}\hfill
\begin{minipage}[c]{0.25\textwidth}
\begin{xmg}
class il {
<morpho> {
 morph <- "il";
 lemma <- "il";
 cat <- cl }}
class ouvre {
<morpho> {
 morph <- "ouvre";
 lemma <- "ouvrir";
 cat   <- v }}
\end{xmg}

%% file: mwes.tex

XMG offers elegant, fully formalized and powerful factoring mechanisms, which enable largely non-redundant grammar engineering. 
However, FrenchTAG, which is one of its most advanced use cases, 
only encodes a limited number of MWE types. Prominently, it covers some inherently reflexive verbs (IRVs), such as \exlitidio{\lex{se} \lex{taire}}{self ignore}{refrain from talking}, in which the reflexive clitic \exlit{se}{self} either is inherent to the verb (i.e. the verb never occurs alone), or markedly changes the meaning and/or the subcategorization frame of the verb. Although the clitic in an IRV occupies the syntactic place of a direct object, it is not a semantic argument of the verb and does not alternate with non-reflexive objects, unlike for transitive verbs. For instance, the 
verb \exlit{ouvrir}{open} assigned to \texttt{n0Nn1} from Fig.~\ref{fig:ftag-classes} exhibits various object realizations  
including a reflexive one (\exlit{Jean s'ouvre}{Jean opens himself}), the latter covered by a combination of tree fragments (d) and (e)  from Fig.~\ref{fig:ftag-tree-fragments}. Conversely, IRVs are assigned to the \texttt{n0ClV} class (cf. Fig.~\ref{fig:ftag-classes}) describing the tree fragment (f), which directly integrates the clitic reflexive \texttt{CL}.

This mechanism displays one possibility of representing restrictive properties 
of MWEs: new classes are created by combining only those alternatives which are allowed by a given type of MWEs. Here, class \texttt{n0ClV} directly combines a free subject with a verb taking a clitic but not any other form of object. 
In early stages of this work, we experimented with this approach.
New classes were created for sets of MWEs which had the same syntactic structure and allowed exactly the same syntactic alternations of their verbal arguments. While some redundancy could be avoided due to factorization of the most common alternations, the number of classes grew rapidly with the addition of new MWEs to the lexicon.

Here, we describe another approach which offers a much higher degree of factorization due to an intensive use of \emph{interface filters} (or \emph{interface FSs}) both in the grammar and in the lexicon. Consider the VMWE in (\ref{ex:prendre-porte}). On the one hand, it shares some properties with transitive verbs (class \texttt{n0Vn1}). For instance, its verb inflects freely, and its subject is unconstrained (canonical, clitic, 
etc.) and agrees with the verb as shown in  (\ref{ex:prendre-porte})--(\ref{ex:prendre-porte-verb-infl}). On the other hand, it exhibits restrictive properties. Its verb cannot be passivized (\ref{ex:prendre-porte-pass}). Its object is lexicalized (\ref{ex:prendre-sortie}) and cannot be cliticized (\ref{ex:prendre-porte-clit}), extracted (\ref{ex:prendre-grande-extr}) or modified (\ref{ex:prendre-grande-porte}).


\begin{examples}
\vspace{-1.5mm}\item \label{ex:prendre-porte} \exlitidio{Jean \lex{prend la porte}}{Jean takes the door}{Jean leaves (because he is forced to)}
\vspace{-3mm}\item \label{ex:prendre-porte-subj-free} \exlitidio{Jean/il \lex{prend la porte}}{Jean/he takes the door}{Jean/he leaves}
\vspace{-3mm}\item \label{ex:prendre-porte-rel-subj} \exlitidio{Jean qui \lex{prend la porte}}{Jean, who takes the door}{Jean, who leaves \ldots}
\vspace{-3mm}\item \label{ex:prendre-porte-verb-infl} \exlitidio{\lex{Prenez la porte}!}{Take.\textsc{2.pl} the door!}{Leave!}
\vspace{-3mm}\item \label{ex:prendre-porte-pass} \#\exlit{La porte est prise par Jean}{The door is taken by Jean}
\vspace{-3mm}\item \label{ex:prendre-sortie} \#\exlit{Jean prend la sortie}{Jean takes the exit}
\vspace{-3mm}\item \label{ex:prendre-porte-clit} \#\exlit{Jean la prend}{Jean takes it}
\vspace{-3mm}\item \label{ex:prendre-grande-extr} \#\exlit{La porte que Jean prend}{The door that Jean takes}
\vspace{-3mm}\item \label{ex:prendre-grande-porte} \#\exlit{Jean prend la grande porte}{Jean takes the big door}
\vspace{-1.5mm}
\end{examples}


\noindent In order to account for these properties, the MWE \ile{\lex{prendre la porte}} receives the lexical class \texttt{mweLemmePrendreLaPorte} from the left column of Listing \ref{fig:mwe-classes}. 
The \texttt{entry} holds the head of the MWE, here \exlit{prendre}{take}, which anchors (in some inflected form) tree templates from the \texttt{mwen0Vn1} family specified in \texttt{fam}. 
A bunch of \texttt{filter} feature-value pairs help to make a selection from the tree templates in \texttt{mwen0Vn1} by unifying with their interface features. For example, the filter \texttt{dia=active} selects the trees that correspond to active diathesis (see also below). The other lexicalized components of \ile{\lex{prendre la porte}} are specified in \texttt{coanchor} where \texttt{ObjectDetNode} and \texttt{ObjNode} are node names from the trees in \texttt{mwen0Vn1}. These names are also used in the \texttt{equation} part in order to pass on morphological features onto the referred nodes.


\begin{figure}[ht]
\input{mwe-classes.tex}
\captionof{lstlisting}{MWE-aware classes based on those from Fig.~\ref{fig:ftag-classes} and decorated with interface constraints}
\label{fig:mwe-classes}
\end{figure}

The right column of Listing \ref{fig:mwe-classes} shows the MWE-aware versions of some FrenchTAG classes from Fig.~\ref{fig:ftag-classes}, in which each syntactic alternation receives an interface constraint, e.g. \texttt{dia=active} in line 12. MWEs having a structure of transitive verbs, like (\ref{ex:prendre-porte}), are assigned to class \texttt{mwen0Vn1}, which includes similar diatheses to \texttt{n0Vn1}. These are however filtered by the MWE lexicon entries: here, the \texttt{dia} filter in the \texttt{mweLemmePrendreLaPorte} class will only admit the \texttt{active} alternation from \texttt{mwen0Vn1}, while all others will be neglected during grammar anchoring prior to parsing. Similarly, a verbal argument in a MWE, here a \texttt{mweSubject} or a \texttt{mweObject}, can be either \texttt{free} or \texttt{lex}icalized. In the former case, the pre-existing FrenchTAG classes \texttt{Subject} and \texttt{Object} are re-used. In the latter, new classes from Fig.~\ref{fig:mwe-lex-object-classes} and \ref{fig:mwe-tree-fragments} are needed, mostly because our objective so far is to keep the original classes intact. 

\begin{figure}[ht]
\input{mwe-lex-object-classes.tex}
\caption{(Simplified) MWE-aware classes describing lexicalized objects of various syntactic structures.}
\label{fig:mwe-lex-object-classes}
\end{figure}

\begin{figure}[ht]
\centering
\setlength{\tabcolsep}{1.4mm}
\begin{tabular}{rlrlrlrl|rc}
(a) & \includegraphics[scale=1]{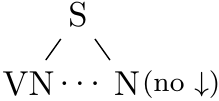} &
(b) & \includegraphics[scale=1]{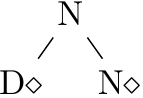} &
(c) & \includegraphics[scale=1]{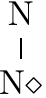} &
(d) & \includegraphics[scale=1]{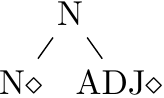} &
(e) & \includegraphics[scale=1]{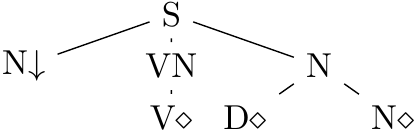} 
\end{tabular}
\caption{Tree fragments defined by the MWE-aware classes from Fig.~\ref{fig:mwe-lex-object-classes}: a canonical object with no substitution node (a), a lexicalized nominal phrase with a fixed \emph{Det-Noun}, \emph{Noun} and \emph{Noun-Adj} structure (b)--(d) --  and the LTAG tree (e) complied from (a), (b) as well as Fig.~\ref{fig:ftag-tree-fragments} (a) and (e). }
\label{fig:mwe-tree-fragments}
\end{figure}

The \texttt{mweObjectLexStruct} class is an alternative of 
3 classes which describe lexicalized noun phrases of the \emph{Det-Noun}, \emph{Noun} or \emph{Noun-Adj} structures, respectively, as in (\ref{ex:prendre-porte}), \exlitidio{Jean \lex{fait \underline{face}} au problème}{Jean makes face to the problem}{Jean deals with the problem} and \exlitidio{Jean \lex{fait \underline{profil bas}}}{Jean makes low profile}{Jean keeps quiet and discreet}. The corresponding tree fragments are depicted in Fig.~\ref{fig:mwe-tree-fragments} (b)--(d). The \texttt{mweObjectLex} class, inherited by \texttt{mweObjectLexStruct}, is similar to \texttt{Object} in Fig.~\ref{fig:ftag-classes} up to substitution marking of the object nodes. For instance, one of the diatheses described by \texttt{mweObjectLex} is the one in Fig.~\ref{fig:mwe-tree-fragments}(a), which is identical to Fig.~\ref{fig:ftag-tree-fragments}(b) up to the substitution mark of the \texttt{N} node. This allows us to unify precisely this node with the root (\texttt{xLexNPTop}) of one of the trees in Fig.\ref{fig:mwe-tree-fragments}(b)--(d), so as to ensure that the object is lexicalized. Note that the \texttt{ObjDetNode} and \texttt{ObjNode} node names, as well as the \texttt{objstruct} filters in Fig.~\ref{fig:mwe-lex-object-classes}, coincide with the lexicon entry 
in Listing \ref{fig:mwe-classes}. This ensures that the filters of this MWE select only the right syntactic alternations, and that its coanchors attach to nodes \texttt{D} and \texttt{N} in Fig.~\ref{fig:mwe-tree-fragments}(e).


\begin{figure}
\includegraphics[scale=0.75]{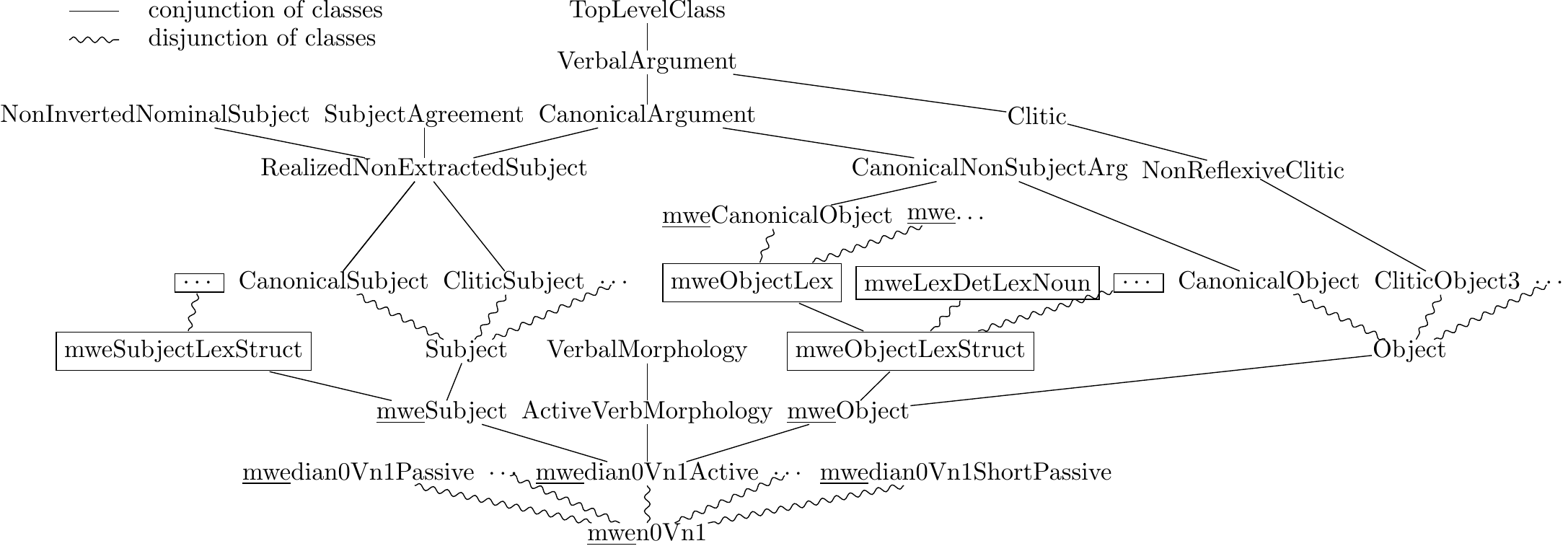} 
\caption{Extract of the XMG class hierarchy in the metagrammar with encoded MWEs. Classes encoding the original families with additional interface filters (and possibly with restricted alternatives) are marked with underlined \underline{mwe} prefixes. New classes encoding the syntactic structure of lexicalized verbal arguments of MWEs are boxed. All other classes remain unchanged with respect to Fig.~\ref{fig:hierarchy-1}.}
\label{fig:hierarchy-2}
\end{figure}
In brief, making FrenchTAG MWE-aware is based on 3 principles: (i) creating MWE lexicon entries with coanchors and interface filters constraining the syntactic alternations, (ii)
reusing pre-existing classes for (more) regular properties and decorating some of them with interface filters, (iii) creating new classes for lexicalized arguments of various
syntactic structures. Fig.~\ref{fig:hierarchy-2} shows an extract of the class hierarchy with the pre-existing classes from Fig.~\ref{fig:ftag-classes}, as well as decorated or newly created MWE-aware classes. Henceforth, we refer to this MWE-aware version of FrenchTAG as mweFrenchTAG.


%% file: mwe-classes.tex
\begin{minipage}[c]{0.42\textwidth}
  \begin{xmg}
class mweLemmePrendreLaPorte {
<lemma> {
 entry <- "prendre";
 cat   <- v;
 fam   <- mwen0Vn1;
 filter dia = active;
 filter subj = free;
 filter obj = lex;
 filter objtype = canonical;
 filter objstruct = lexDetLexN;
 coanchor ObjDetNode -> "la"/d;
 coanchor ObjNode -> "porte"/n;
 equation ObjNode -> gen=f;
 equation ObjNode -> num=sg }}
\end{xmg}
\end{minipage}\hfill
\begin{minipage}[c]{0.59\textwidth}
\begin{xmg}
class mweSubject {
 Subject[] *= [subj=free]
 | mweSubjectLexStruct[] *= [subj=lex]}
class mweObject {
 Object[] *= [obj=free]
 | mweObjectLexStruct[] *= [obj=lex]}
class mwedian0Vn1Active{
 mweSubject[]; 
 activeVerbMorphology[];
 mweObject[] }
class mwen0Vn1 {
 mwedian0Vn1Active[] *= [dia=active]
 |mwedian0Vn1Passive[] *= [dia=passfull]
 |mwedian0Vn1ShortPassive[] *= [dia=passshort]}
\end{xmg}
\end{minipage}

%% file: mwe-lex-object-classes.tex
\begin{xmg}
class mweObjectLexStruct
 import	mweObjectLex[]
 declare ?lexNP {
 {   ?lexNP = mweLexDetLexNoun[ObjDetNode,ObjNode] *= [objstruct=lexDetLexN]
   | ?lexNP = mweNoDetLexNoun[ObjNode] *= [objstruct=lexN]
   | ?lexNP = mweLexNounLexAdj[ObjNode,ObjAdjNode] *= [objstruct=lexNLexAdj] };
 ?LexObj = ?lexNP.xLexNPTop }
\end{xmg}


%% file: eval.tex

This work is meant to provide a proof of concept that non-redundant lexical encoding of MWEs can be effectively achieved in XMG. Therefore, we do not aim at evaluating the coverage of the metagrammar. Instead, we wish to see: (i) to what extent the metagrammar allows us to cover the properties of MWEs of a variable degree of regularity, as they are present in real data, (ii) how far the metagrammar changes or extends, when new MWEs and phenomena are to be covered. 

To this end, we used the French part of the PARSEME corpus of VMWEs \cite{Savaryatal:forth}. 
For the $1,449$ unique VMWEs, their idiomatic and some literal occurrences 
were extracted. 
From the resulting $5,893$ 
occurrences an evaluation dataset was
selected following three criteria: frequency, syntactic variety, and presence
of literal readings.
%
We first chose 14 most frequent VMWE of various categories 
and checked that some of them exhibited literal readings in the dataset.
Their total number of occurrences was equal to 1,004. We needed a
dataset of a more manageable size but representing a
large variety of syntactic phenomena. To this end, each occurrence was abstracted as a sequence of components, each component represented by (i) its POS, part-of-speech tag, and (ii) the set of outgoing dependency relation types limited to 2 (first two in the lexicographic order). This reduced the 1,004 occurrences to 225 classes, from which we repeatedly randomly selected a subset of 56 occurrences (4 for each of the 14 VMWEs). The syntactic variety, in terms of Shannon's entropy, of VMWE occurrences inside each subset was calculated. The subset maximizing the entropy was then randomly divided into two disjoint subsets: DEV (50\%) and TEST (50\%).


We initially intended to use literal readings to demonstrate the way the grammar represents ambiguous MWEs. However, it turned out that most of them were embedded in other MWEs.
We therefore abandoned them in this study and thus reduced DEV and TEST to 26 
sentences each. 

The original FrenchTAG is focused on verbal predicates and their arguments, while providing only basic coverage for other syntactic categories. It also has a rather scarce lexicon. 
Therefore, the 52 sentences were manually simplified so as to: (i) avoid subordinate sentences and coordinations, (ii) reduce the non-lexicalized arguments of the verbs (to the head nouns, or to single-word proper names and adverbials). 
The simplified datasets, DEV-S and TEST-S, are cited in the Appendix.\footnote{The original and simplified datasets, and the simplification rules, are available online at: ANONYMIZED URL}
%
The lexicon necessary for parsing them was automatically derived from the Unitex corpus processor.\footnote{http://unitexgramlab.org/}
Then, in phase 1, most MWEs occurring in DEV-S were encoded in mweFrenchTAG (additionally to a dozen of previously encoded examples), so as to parse at least the syntactic alternations occurring in DEV-S. As shown in Tab.~\ref{tab:results}, after this phase, we observed an 18\% increase in the number of XMG classes covering the grammar (i.e. not the lexicon), which is substantial but expected,  since this phase included re-engineering the class hierarchy to make it MWE-ware (cf. Sec.~\ref{sec:mwes}). In phase 2, we repeated the same actions for TEST-S. We noted only a 1.1\% increase in the number of classes between phases 1 and 2. We expect this increase to be even lower in the next encoding iterations, as new MWEs should exhibit increasingly similar syntactic structures and properties to those already encoded. Experimental large-scale validation of 
this hypothesis is 
left to future work.

\begin{table}[ht]
\centering
\scalebox{0.9}{
\begin{tabular}{|l|r|r|r|r|}
\hline
\multirow{2}{*}{} & \multirow{2}{*}{\textbf{FrenchTAG}} & \multicolumn{2}{|c|}{\textbf{mweFrenchTAG with MWEs from}} \\
                  &                               &  \multicolumn{1}{|c|}{\textbf{DEV-S}} & \multicolumn{1}{|c|}{\textbf{DEV-S + TEST-S}} \\\hline
classes    & 285 & 337 (+18\%) & 341 (+1.1\%) \\\hline
MWE lemmas & 5  & 31 (+520\%) & 37 (+19\%) \\ \hline
\end{tabular}}
\caption{The growth of FrenchTAG as a result making it MWE-aware. }
\label{tab:results}
\end{table}

Some examples in DEV-S and TEST-S have not been successfully encoded so far. Example (\ref{ex:dev:mis-en-examen}) 
contains a regular extraction whose type seems not to be covered by FrenchTAG. The MWEs in (\ref{ex:dev:se-faire-adj}) and (\ref{ex:test:se-trouver-adj}) are reflexive but have a behavior of a copula 
and their optimal representation requires more insight. Finally, the encoding of the MWE in (\ref{ex:dev:porter-nom-1}), (\ref{ex:dev:porter-nom-2}), (\ref{ex:test:porter-nom-1}) and (\ref{ex:test:porter-nom-2}) is partly satisfactory since the implementation of a compulsory but non-lexicalized modifier has not yet been solved in our approach. Note that the ratio of not (or partly) encoded
MWEs decreases between phases 1 and 2.

%% file: conc.tex
There is a fundamental tension between the flexibility of MWEs, that is their unforeseeable but possibly recurring ``irregularities'', and lexical encoding systems which ought to be denotationally precise and at the same time maximally theory-independent. In this paper, we have argued that XMG incorporates both the necessary flexibility and denotational rigor. To this end, we have shown how to extend the metagrammar underlying FrenchTAG, a large-coverage grammar of French, by grammar-specific classes to which the rather grammar-agnostic descriptions of MWEs can be linked.  By evaluation against a MWE-annotated reference corpus, we noticed a considerable drop in the growth of grammar size during incremental implementation, which seems to suggest that the redundancy of MWE is efficiently captured.  

There are several work packages left for future work: (i) to extend the approach to challenging phenomena such as co-indexation; (ii) to make the XMG compiler and LTAG parser also handle compulsory or prohibited modification of anchor or co-anchor nodes; (iii) to allow for co-anchoring lemmas rather than fixed inflected forms; (iv) to introduce intermediate classes into the original hierarchy from Fig.~\ref{fig:ftag-lemmas} which directly account for more co-anchoring phenomena.\footnote{These classes should represent tree fragments in which non-terminal leaves can either be substitution nodes or unifiable with subtrees representing lexicalized arguments. In this way, redundancy could be avoided e.g. between tree fragments from Fig.~\ref{fig:ftag-tree-fragments}(b) and Fig.~\ref{fig:mwe-tree-fragments} (a).} As to the last point,  our strategy was to keep the original FrenchTAG classes intact, but this is sub-optimal in cases where co-anchoring becomes massive. Finally, what needs to be done in the long run is to investigate how the MWE descriptions we propose can be reused with grammars from other frameworks, for example HPSG or LFG.


